\documentclass{article}

\usepackage{graphicx} 
\usepackage{subfigure} 

\usepackage{natbib}

\usepackage{algorithm}
\usepackage{algorithmic}

\usepackage{hyperref}

\usepackage{amsmath}
\usepackage{amssymb}

\usepackage{tabularx}
\newcolumntype{D}[1]{>{\centering}p{#1}}

\usepackage[accepted]{icml2011} 

\icmltitlerunning{Functional Regularized Least Squares Classification with Operator-valued Kernels}

\begin{document} 

\twocolumn[
\icmltitle{Functional Regularized Least Squares Classification 
           with Operator-valued Kernels}

\icmlauthor{Hachem Kadri}{hachem.kadri@inria.fr}
\icmladdress{SequeL Project, INRIA Lille - Nord Europe, Villeneuve d'Ascq, France}
\icmlauthor{Asma Rabaoui}{asma.rabaoui@ims-bordeaux.fr}
\icmladdress{LAPS-IMS/CNRS, Universit\'e de Bordeaux, Talence, France}
\icmlauthor{Philippe Preux}{philippe.preux@inria.fr}
\icmladdress{LIFL/CNRS/INRIA, Universit\'e de Lille, Villeneuve d'Ascq, France}
\icmlauthor{Emmanuel Duflos}{emmanuel.duflos@ec-lille.fr}
\icmladdress{LAGIS/CNRS/INRIA, Ecole Centrale de Lille, Villeneuve d'Ascq, France}
\icmlauthor{Alain Rakotomamonjy}{alain.rakotomamonjy@insa-rouen.fr}
\icmladdress{LITIS, UFR de Sciences, Universit\'e de Rouen, St Etienne du Rouvray, France}

\icmlkeywords{operator-valued kernels, functional classification, sound recognition}

\vskip 0.3in
]

\begin{abstract} 
Although operator-valued kernels have recently received increasing interest in various
machine learning and functional data analysis problems such as multi-task learning or 
functional regression, little attention has been paid to the understanding of their associated 
feature spaces. In this paper, we explore the potential of adopting an operator-valued kernel 
feature space perspective for the analysis of functional data. We then extend the 
Regularized Least Squares Classification (RLSC) algorithm to cover situations where 
there are multiple functions per observation. 
Experiments on a sound recognition problem show that the proposed method outperforms 
the classical RLSC algorithm.

\end{abstract}

\section{Introduction}
Following the development of multi-task and complex output learning methods~\cite{OpenHouse-2006}, operator-valued 
kernels have recently attracted considerable attention in the machine learning 
community~\cite{Micchelli-2005c, Reisert-2007, Caponnetto-2008}.~It turns out that these kernels lead 
to a new class of algorithms well suited for learning multi-output functions. For example, in multi-task learning contexts, 
standard single-task kernel learning methods such as support vector machines~(SVM) 
and regularization networks~(RN) are extended to deal with several dependent tasks at once by learning a vector-valued 
function using multi-task kernels~\cite{Evgeniou-2005}.~Also, in functional data analysis~(FDA) where 
observed continuous data are measured over a densely sampled grid and then represented by real-valued functions 
rather than by discrete finite dimensional vectors, function-valued reproducing kernel Hilbert spaces~(RKHS) are constructed 
from nonnegative operator-valued kernels to extend kernel ridge regression from finite dimensions to the infinite 
dimensional case~\cite{Lian-2007, kadri-2010}. 

While most recent work has focused on studying operator-valued kernels and their corresponding RKHS from the perspective 
of extending Aronszajn's pioneering work~\yrcite{Aronszajn-1950} to the vector or function-valued 
case~\cite{Micchelli-2005a, Carmeli-2010, kadri-2010}, in this paper we pay special 
attention to the feature space point of view~\cite{Sch-1999}. More precisely, we provide some ideas targeted at 
advancing the understanding of feature spaces associated with operator-valued kernels and we show how these kernels 
can design more suitable feature maps than those associated with scalar-valued kernels, especially when input data are 
complex, infinite dimensional objects like curves and distributions. In many experiments, the observations consist of a 
sample of random functions or curves. So, we adopt in this paper a functional data analysis point of view in which each curve 
corresponds to one observation. This is an extension of multivariate data analysis where observations consist of vectors of 
finite dimension. For an introduction to the field of FDA, the two monographs by Ramsay \& 
Silverman~\yrcite{Ramsay-2005, Ramsay-2002} provide a rewarding and accessible overview on foundations and applications, as
well as a collection of motivating examples~\cite{Muller-2005}. 

To explore the potential of adopting an operator-valued kernel feature space approach, we are interested in the problem of 
functional classification in the case where there are multiple functions per observation.~This study is valuable from a 
variety of perspectives.~Our motivating example is the practical problem of sound recognition which is of great importance in 
surveillance and security applications~\cite{Dufaux-2000, Istrate-2006, Asma-2008}. 
This problem can be tackled by classifying incoming signals representing 
the environmental sounds into various predefined classes. In this setting, a preprocessing step consists in applying 
signal-processing techniques to generate a set of features characterizing the signal to be classified. These features form a 
so-called feature vector which contains discrete values of different functional parameters providing information about 
temporal, frequential, cepstral and energy characteristics of the signal. In standard machine learning methods, the feature 
vector is considered to be a subset of $\mathbb{R}^n$ by concatenating samples of the different functional features, 
and this has the drawback of not considering any dependencies 
between different values over subsequent time-points within the same functional datum.~Employing these methods implies that 
permuting time points arbitrarily, which is equivalent to exchanging the order of the indexes in a 
multivariate vector, should not change the result of statistical analysis. Taking into account the inherent sequential 
nature of the data and using the dependencies along the time-axis should lead to higher quality results~\cite{lee-2004}. 
In our work, we use a functional data analysis approach based on modeling each sound signal by a vector of functions 
(in $(L^2)^p$ for example, where $L^2$ is the space of square integrable functions and $p$ is the number of functional 
parameters) 
rather than by a vector 
in $\mathbb{R}^n$, with the hope of improving performance by: (1) considering the relationship between samples of a 
function variable and thus the dynamic behavior of the functional data, (2) capturing discriminative characteristics 
between functions contrary to the concatenation procedure.

During the last decade, various kernel-based methods have become very popular for solving classification problems. Among 
them, the regularized least squares classification (RLSC) is a simple regularization algorithm which achieves good 
performance, nearly equivalent to that of the well-known Support Vector Machines 
(SVMs)~\cite{Rifkin-2003,Rifkin-2007,Rifkin-2004,Zhang-2004}. By using operator-valued kernels, we extend the RLSC algorithm to 
cover situations where there are multiple functions per observation. One main obstacle for this extension involves the 
inversion of the block operator kernel matrix (kernel matrix where the block entries are linear operators). 
In contrast to the situation in the multivariate case, this inversion is not always feasible in infinite dimensional 
Hilbert spaces. In this paper, we attempt to overcome this problem by characterizing a class of operator-valued kernels 
and performing an eigenvalue decomposition of this kernel matrix. 

The remainder of this paper is organized as follows. In section~\ref{fs}, we review concepts of operator-valued kernels 
and their corresponding reproducing kernel Hilbert spaces and discuss some ideas for understanding the associated feature 
maps. Using these kernels, we propose a functional regularized least squares classification algorithm in 
section~\ref{frlsc}. It is an extension of the classical RLSC to the case where there are multiple functions per observation. 
The proposed algorithm is experimentally evaluated on a sound recognition task in section~\ref{exp}. Finally, section~\ref{conc} 
presents some conclusions and future work directions.

\section{Operator-valued kernels and associated feature spaces}
\label{fs}
In the machine learning literature, operator-valued kernels were first introduced by Micchelli and 
Pontil~\yrcite{Micchelli-2005c} to deal with the problem of multi-task learning. In this context, these kernels, called 
multi-task kernels, are matrix-valued functions and their corresponding vector-valued reproducing kernel Hilbert spaces are 
used to learn multiple tasks simultaneously~\cite{Evgeniou-2005}.~Most often, multi-task kernels are constructed from 
scalar-valued kernels which are carried over to the vector-valued setting by a positive definite matrix. More details and 
some examples of multi-task kernels can be found in~\cite{Caponnetto-2008}. 
In addition, some works have studied the construction of such kernels from a functional data analysis point of view 
(continuous data). For example, in~\cite{kadri-2010}, the authors showed how infinite dimensional 
operator-valued kernels can be used to perform nonlinear functional regression in the case where covariates as well as 
responses are functions.

Since we are interested in the problem of functional classification, we use a similar framework to that 
of~\cite{kadri-2010}, but we consider the more general case where we have multiple functions as inputs to the 
learning module. Kernel-based learning methodology can be extended directly from the vector-valued to the functional-valued 
case.
The principle of this extension is to replace vectors by functions and matrices by linear operators; 
scalar products in vector space are replaced by scalar products in function space, which is usually chosen as the space of
square integrable functions $L^2$ on a suitable domain. 
In the present paper, we focus on infinite dimensional operator-valued kernels and their corresponding 
functional-valued RKHS. 

An operator-valued kernel $K : X \times X \longrightarrow \mathcal{L}(Y)$ is the reproducing kernel of a Hilbert space of 
functions from an input space $X$ which takes values in a Hilbert space $Y$. $\mathcal{L}(Y)$ is the set of all bounded 
linear operators from $Y$ into itself. Input data are represented by a vector of functions, so we consider the case where 
$X \subset (L^2)^p$ and $Y \subset L^2$. Function-valued RKHS theory is based on the \textit{one-to-one correspondence} 
between 
reproducing kernel Hilbert spaces of function-valued functions and positive operator-valued kernels. We start by recalling 
some basic properties of such spaces. We say that a Hilbert space $\mathcal{F}$ of functions 
$f: X\longrightarrow Y$ has the \textit{reproducing property}, if $\forall x\in X$ the linear functional 
$f\longrightarrow \langle f(x),y\rangle_Y$ is continuous for any $x\in X$ and $y\in Y$.  By the Riesz representation 
theorem it follows that for a given $x\in X$ and for any choice of $y\in Y$, there exists an element 
$h_x^y \in \mathcal{F}$, s.t.
\begin{equation*}
 \forall f \in \mathcal{F}\ \ \ \langle h_x^y,f\rangle_{\mathcal{F}} = \langle f(x),y\rangle_Y
\end{equation*}
We can therefore define the corresponding operator-valued kernel $K: X \times X \longrightarrow \mathcal{L}(Y)$ such that
\begin{equation*}
 \langle K(x_1,x_2)y_1,y_2\rangle_Y = \langle h_{x_1}^{y_1},h_{x_2}^{y_2} \rangle_{\mathcal{F}}
\end{equation*}
It follows that 
\begin{equation*}
 \langle h_{x_1}^{y_1}(x_2), y_2\rangle_Y = \langle h_{x_1}^{y_1},h_{x_2}^{y_2} \rangle_{\mathcal{F}} 
= \langle K(x_1,x_2)y_1,y_2\rangle_{\mathcal{G}_{y}}
\end{equation*}
and thus we obtain the reproducing property
\begin{equation}
\label{reproducing}
 \langle K(x,.)y,f\rangle_{\mathcal{F}} = \langle f(x),y\rangle_Y
\end{equation}

Consequently, we obtain that $K(.,.)$ is a positive definite operator-valued kernel as defined below: 
(see proposition 1 in~\cite{Micchelli-2005c} for the proof)

\textbf{Definition:} We say that $K(x_1,x_2)$, satisfying $K(x_1,x_2) = K(x_2,x_1)^*$ 
(the superscript $*$ indicates the adjoint operator), is a positive definite 
operator-valued kernel
if given an arbitrary finite set of points $\{(x_i,y_i)\}_{i=1,\ldots,n}\in X\times Y$, the corresponding 
block matrix $K$ with $K_{ij} = \langle K(x_i,x_j)y_i, y_j \rangle_Y$ is positive semi-definite. 

Importantly, the converse is also true. Any positive operator-valued kernel $K(x_1,x_2)$ gives rise
to an RKHS $\mathcal{F}_K$, which can be constructed by considering the space of function-valued functions $f$ having the 
form $f(.) = \sum_{i=1}^n K(x_i,.)y_i$ and taking completion with respect to the inner product given by 
$\langle K(x_1,.)y_1, K(x_2,.)y_2 \rangle_{\mathcal{F}} = \langle K(x_1,x_2)y_1,y_2 \rangle_Y$.

In the following, we present an example of function-valued RKHS with functional inputs and the associated 
operator-valued kernel. 

\textbf{Example.} Let $X=H$ and $Y = L^2(\Omega)$, where $H$ is the Hilbert space of constants in $[0,1]$
and $L^2(\Omega)$ the space of square integrable functions on $\Omega$. We denote by $\mathcal{M}$ 
the space of $L^2(\Omega)$-valued functions on $H$ whose norm 
$ \|g\|_{\mathcal{M}}^2 = \displaystyle\int_\Omega \int_H [g(v)(x)]^2 dv dx$ is finite.

Let $(\mathcal{F};\langle .,. \rangle_{\mathcal{F}})$ be the space of functions from $H$ to $L^2(\Omega)$ such that:
\begin{equation*}
  \left\{
    \begin{split}
    & \mathcal{F} = \{f,\ \exists f' = \displaystyle\frac{d f(v)}{d v} \in \mathcal{M}, 
f(u) = \int_0^u f'(v) dv \}\\ 
   & \langle f_1, f_2 \rangle_\mathcal{F} = \langle f'_1, f'_2 \rangle_{\mathcal{M}}
    \end{split}
  \right.
\end{equation*}
$\mathcal{F}$ is a RKHS with kernel $K(u,v) = M_{\varphi(u,v)}$. $M_\varphi$ is the multiplication operator associated 
with the function $\varphi$ where $\varphi(u,v)$ is equal to $u$ if $u(x) \leq v(x)\ \forall x \in \Omega$ and $v$ otherwise. 
It is easy to check that $K$ is Hermitian and nonnegative. Now we show that the reproducing property holds for any $f\in \mathcal{F}$, 
$w\in L^2(\Omega)$ and $u\in H$
\begin{align*}
\begin{array}{l}
  \langle f, K(u,.)w \rangle_{\mathcal{F}} = \langle f', [K(u,.)w]' \rangle_{\mathcal{M}}\\[0.1cm]
= \displaystyle\int_\Omega \int_H  [f'(v)](x) [K(u,v)w]'(x) dv dx \\[0.3cm]
= \displaystyle\int_\Omega \int_0^u  [f'(v)](x) w(x) dv dx 
= \displaystyle\int_\Omega  [f(u)](x) w(x) dx \\[0.3cm]
=  \langle f(u), w \rangle_{L^2(\Omega)} \hfill \blacksquare
\end{array}
\end{align*}

Similar to the scalar case, operator-valued kernels provide an elegant way of dealing with nonlinear algorithms by reducing 
them to linear ones in some feature space $F$ nonlinearly related to input space. A feature map associated with 
an operator-valued kernel $K$ is a continuous function 
\begin{equation*}
 \Phi: X \times Y \longrightarrow \mathcal{L}(X,Y)
\end{equation*}
such that, for every $x_1, x_2 \in X$ and $y_1, y_2 \in Y$
\begin{equation*}
 \langle K(x_1,x_2)y_1, y_2 \rangle_{Y} = \langle \Phi(x_1,y_1), \Phi(x_2,y_2) \rangle_{\mathcal{L}(X,Y)}
\end{equation*}
where $\mathcal{L}(X,Y)$ is the set of mappings from $X$ to $Y$. 
By virtue of this property, $\Phi$ is called a \textit{feature map associated with} $K$. 
Furthermore, from~(\ref{reproducing}), it follows that in particular
\begin{equation*}
\langle K(x_1,.)y_1, K(x_2,.)y_2 \rangle_{\mathcal{F}} = \langle K(x_1,x_2)y_1, y_2 \rangle_{Y}
\end{equation*}
which means that any operator-valued kernel admits a feature map representation with a feature space 
$\mathcal{F} \subset \mathcal{L}(X,Y)$, and corresponds to a dot product in another space. 

From this feature map perspective, we study the geometry of a feature space associated with an operator-valued kernel 
and we compare it with the one obtained by a scalar-valued kernel. More precisely, we consider two reproducing kernel 
Hilbert spaces (RKHS) $\mathcal{F}$ and $\mathcal{H}$. $\mathcal{F}$ is a RKHS of function-valued functions on $X$ with 
values in $Y$. $X \subset (L^2)^p$, $Y \subset L^2$ and let $K$ be the reproducing operator-valued kernel of $\mathcal{F}$. 
$\mathcal{H}$ is also a RKHS, but of scalar-valued functions on $X$ with values in $\mathbb{R}$, and $k$ its reproducing 
real-valued kernel. The mappings $\Phi_K$ and $\Phi_k$ associated, respectively, with the kernels $K$ and $k$ are defined 
as follows
\begin{equation}
\nonumber
 \Phi_K^y: (L^2)^p \rightarrow \mathcal{L}((L^2)^p,L^2),\quad x \mapsto K(x,.)y
\end{equation}
and
\begin{equation}
\nonumber
 \Phi_k: (L^2)^p \rightarrow \mathcal{L}((L^2)^p,\mathbb{R}),\quad x \mapsto k(x,.)
\end{equation}
These feature maps can be seen as a mapping of the input data $x_i$, which are vector of functions in $(L^2)^p$ , 
into a feature space in which the dot product can be computed using the kernel functions. This idea leads to design nonlinear 
methods based on linear ones in the feature space. In a supervised classification problem for example, 
since kernels could map input data into a higher dimensional space, 
kernel methods deal with this 
problem by finding a linear separation in the feature space between data which can not be separated linearly 
in the input space. 
We now compare the dimension of feature spaces obtained by the maps $\Phi_K$ and $\Phi_k$. To do this, we adopt a functional 
data analysis point of view where observations are composed of sets of functions. 
Direct understanding of this FDA viewpoint comes from the consideration of the ``atom'' of a statistical analysis. 
In a basic course in statistics, atoms are ``numbers'', while in multivariate data analysis the atoms are vectors and methods 
for understanding populations of vectors are the focus. FDA can be viewed as the generalization of this, where the atoms are 
more complicated objects, such as curves, images or shapes represented by functions~\cite{Zhao-2004}. 
Based on this, the dimension of the input space is $p$ since $x_i\in (L^2)^p$ is a vector of $p$ functions. The feature 
space obtained by the map $\Phi_k$ is a space of functions, so its dimension 
from a FDA point of view is one. The map $\Phi_K$ projects the input data 
into a space of operators $\mathcal{L}(X,Y)$. This means that using the operator-valued kernel $K$ corresponds to mapping 
the functional data $x_i$ into a higher, possibly infinite, dimensional space $(L^2)^d$ with $d\rightarrow \infty$. 
In a binary functional classification problem, we have higher probability to achieve linear separation between the classes by 
projecting the functional data into a higher dimensional feature space rather than into a lower one, that is why we think 
that it is more suitable to use operator-valued than scalar-valued kernels in this context.

\section{Functional regularized least squares classification}
\label{frlsc}

In this section, we show how to extend the regularized least squares classification algorithm (RLSC) to functional 
contexts using operator-valued kernels. 
To use these kernels for a classification problem, we consider the labels to be functions in some function space rather 
than real values as usual. The functional classification problem can then be framed as that of learning a function-valued 
function $f: X \longrightarrow Y$ where $X \subset (L^2)^p$ and $Y \subset L^2$. The RLSC algorithm~\cite{Rifkin-2003} 
is based on solving a Tikhonov minimization problem associated with a square loss function, and then an estimate 
$f^*$ of $f$ in a Hilbert space $\mathcal{F}$ with reproducing operator-valued kernel 
$K:X\times X \longrightarrow \mathcal{L}(Y)$ is obtained by minimizing
\begin{equation}
\label{mp}
f^* = \arg\min\limits_{f \in \mathcal{F}}\sum\limits_{i=1}^{n}\|y_{i}-f(x_{i})\|_{Y}^{2}
+\lambda\|f\|_{\mathcal{F}}^{2}
\end{equation}
By the representer theorem~\cite{Micchelli-2005a,kadri-2010}, the solution of this problem has the following form 
\begin{equation}
\label{llll}
 f^*(x) = \sum_{j=1}^n K(x,x_j)\beta_j, \quad \beta_j\in Y
\end{equation}
Substituting~(\ref{llll}) in~(\ref{mp}) , we come up with the following minimization over the scalar-valued functions $\beta_i$ 
rather than the function-valued function $f$ 
\begin{eqnarray}
\label{mp1}
\begin{array}{cc}
    \min\limits_{\beta_v\in (Y)^n}
& \sum\limits_{i=1}^{n}\|\beta_{i}-\sum\limits_{j=1}^{n}
K(x_{i},x_{j})\beta_{j}\|_{Y}^{2}  
     \\ & +\lambda
    \sum\limits_{i,j}^{n}\langle K(x_{i},x_{j})\beta_{i},\beta_{j}\rangle_{Y} 
\end{array}
\end{eqnarray}
$\beta_v$ is the vector of functions $(\beta_i)_{i=1,\dots,n} \in (L^2)^n$. 
The problem~(\ref{mp1}) can be solved in three ways. Assuming that the observations are made on a regular grid 
$\{t_1,\ldots,t_m\}$, one can first discretize the functions $x_i$ and $y_i$ and then solve the problem using multivariate 
data analysis techniques. However, this has the drawback, as well known in the FDA literature, 
of not considering the relationships that exist between samples. The second way consists in considering the output space 
$Y$ to be a scalar valued reproducing Hilbert space. In this case, the functions $\beta_i$ can be approximated by a 
linear combination of a scalar kernel $\beta_i = \sum_{l=1}^m\alpha_{il}k(s_l,.) $ and then the problem~(\ref{mp1}) 
becomes a minimization problem over the real values $\alpha_{il}$. Another possible way to solve the minimization~(\ref{mp1}) is to compute its derivative using the directional derivative and setting the result to zero 
to find an analytic solution of the problem. 
It follows that $\beta_v$ satisfies the system of linear operator 
equations
\begin{equation}
\label{sloe}
 (\mathcal{K}+\lambda I)\beta_v = y_v
\end{equation}
where $\mathcal{K} = [K(x_i,x_j)]_{i,j=1}^n$ is a $n\times n$ block operator matrix 
($\mathcal{K}_{ij} \in \mathcal{L}(Y)$) and $y_v$ the vector of functions $(y_i)_{i=1}^n \in (L^2)^n$. 

In this work, we are interested in this third approach which extends the 
classical RLSC algorithm to functional data analysis domain. 
One main obstacle for this extension is the inversion of the block operator kernel matrix $\mathcal{K}$. 
Block operator matrices generalize block matrices to the case where the block entries are linear operators between 
infinite dimensional Hilbert spaces.
In contrast to the situation in the multivariate case, inverting such matrices is not always feasible in infinite 
dimensional spaces. To overcome this problem, we study the eigenvalue decomposition of a class of block operator 
kernel matrices obtained from operator-valued kernels having the following form
\begin{equation}
\label{kern}
 K(x_i,x_j) = G(x_i,x_j)T,\quad \forall x_i,x_j \in X
\end{equation}
where $G$ is a scalar-valued kernel and $T$ is an operator in $\mathcal{L}(Y)$.~This kernel 
construction is adapted from~\cite{Micchelli-2005a,Micchelli-2005c}.~Choosing $T$ depends on the context. For multi-task kernels, $T$ is a finite dimensional matrix which model relations 
between tasks. In FDA, Lian~\yrcite{Lian-2007} suggested the use of the identity operator, while Kadri et al.~\yrcite{kadri-2010}
showed that it will be more useful to choose other operators than identity that are able to take into account 
functional properties of the input and output spaces. They introduced a functional extension of the Gaussian kernel 
based on the multiplication operator. In this work, we are interested in kernels constructed from the integral operator. 
This seems to be a reasonable choice since functional linear model (see Eq.~(\ref{flm})) 
are based on this operator~\cite{Ramsay-2005} 
\begin{equation}
\label{flm}
y(s) = \alpha(s) + \int x(t) \nu(s,t) dt 
\end{equation}
where $\alpha$ and $\nu$ are the functional parameters of the model. 
So we consider the following positive definite operator-valued kernel
\begin{equation}
\label{kernn}
 (K(x_i,x_j)y)(t) = G(x_i,x_j)\int_\Omega e^{-|t-s|} y(s) ds
\end{equation}
where $y\in Y$ and $\{s,t\}\in \Omega =[0,1]$. Note that a similar kernel was proposed in~\cite{Caponnetto-2008} for 
linear spaces of functions from $\mathbb{R}$ to $Y$. 
\begin{algorithm}[tb]
   \caption{Functional RLSC}
   \label{alg:frlsc}
\begin{algorithmic}
   \STATE {\bfseries Input} 
    \\ $\quad$ data $x_i\in (L^2([0,1]))^p$, size $n$
    \\ $\quad$ labels $y_i\in L^2([0,1])$, size $n$
    \\ \textbf{Eigendecomposition of $\mathcal{G}$}
    \\ $\quad$ $\mathcal{G} =  G(x_i,x_j)_{i,j=1}^n \in \mathbb{R}^{n \times n}$
    \\ $\quad$ eigenvalues $\alpha_i \in \mathbb{R}$, size $n$
    \\ $\quad$ eigenvectors $v_i \in \mathbb{R}^n$, size $n$
    \\ \textbf{Eigendecomposition of $T$}
    \\ $\quad$ $T \in \mathcal{L}(Y)$
    \\ $\quad$ Initialize $k$: number of eigenfunctions
    \\ $\quad$ eigenvalues $\delta_i \in \mathbb{R}$, size $k$
    \\ $\quad$ eigenfunctions $w_i \in L^2([0,1])$, size $k$
    \\ \textbf{Eigendecomposition of $\mathcal{K} = \mathcal{G} \otimes T$}
    \\ $\quad$ $\mathcal{K} =  K(x_i,x_j)_{i,j=1}^n \in (\mathcal{L}(Y))^{n \times n}$
    \\ $\quad$ eigenvalues $\theta_i \in \mathbb{R}$, size $n\times k$
    \\ $\quad$ $\quad$ $\quad$ $\theta = \alpha \otimes \delta$
    \\ $\quad$ eigenfunctions $z_i \in (L^2([0,1]))^n$, size $n\times k$
    \\ $\quad$ $\quad$ $\quad$ $z = v \otimes w$
    \\ \textbf{Solution of~(\ref{mp1}) $\beta = (\mathcal{K}+\lambda I)^{-1} y$}
    \\ $\quad$ Initialize $\lambda$: regularization parameter
    \\ $\quad$ $\beta = \sum_{i=1}^{n\times k}  (\theta_i+\lambda)^{-1} \sum_{j=1}^n\langle z_{i,j}, y_j \rangle z_{i}$
\end{algorithmic}
 \end{algorithm}
The $n\times n$ block operator kernel matrix $\mathcal{K}$ of operator-kernels having the form~(\ref{kern}) can be 
expressed by a Kronecker product between the matrix $\mathcal{G} = G(x_i,x_j)_{i,j=1}^n$ in $\mathbb{R}^{n \times n}$ 
and the operator $T\in \mathcal{L}(Y)$
\begin{equation}
\nonumber
 \mathcal{K} = 
\begin{pmatrix}
G(x_1,x_1)T & \ldots & G(x_1,x_n)T \\
\vdots & \ddots & \vdots \\
G(x_n,x_1)T & \ldots & G(x_n,x_n)T \\
\end{pmatrix} 
= \mathcal{G} \otimes T
\end{equation}
In this case, the eigendecomposition of the matrix $\mathcal{K}$ can be obtained from the eigendecompositions of 
$\mathcal{G}$ and $T$ (see Algorithm~\ref{alg:frlsc}). Let $\theta_i$ and $z_i$ be, respectively, the eigenvalues and 
the eigenfunctions of $\mathcal{K}$, the inverse operator $\mathcal{K}^{-1}$ is given by 
\begin{equation}
 \nonumber
\mathcal{K}^{-1}y_v = \sum_i \theta_i^{-1} \langle y_v,z_i \rangle z_i, \quad \forall y_v \in (L^2)^n
\end{equation}
Now we are able to solve the system of linear operator equation~(\ref{sloe}) and the functions $\beta_i$ can be 
computed from eigenvalues and eigenfunctions of the matrix $\mathcal{K}$, as described in Algorithm~\ref{alg:frlsc}.
%
%
%
%
%
%
\section{Experiments}
\label{exp}
Our experiments are based on a sound recognition task. The performance of 
the functional RLSC algorithm, described in section~\ref{frlsc}, is evaluated on a data set of sounds collected from
commercial databases which include sounds ranging from screams to explosions, such as gun shots or glass breaking, 
and compared with the RLSC method~\cite{Rifkin-2003}.

Many previous works in the context of sound recognition have concentrated on classifying environmental
sounds other than speech and music~\cite{Dufaux-2000,Peltonen-ICASSP-2002}. Such
sounds are extremely versatile, including signals generated in domestic, business, and outdoor environments. 
A system that is able to recognize such sounds may be of great importance for surveillance
and security applications~\cite{Istrate-2006,Asma-2008}. 
The classification of a sound is usually performed in two steps.
First, a pre-processor applies signal processing techniques to
generate a set of features characterizing the signal to be
classified. Then, in the feature space, a decision rule is implemented to assign a class to a pattern.

\subsection{Database description}
As in~\cite{Asma-2008}, the major part of the sound samples used in the recognition
experiments is taken from two sound libraries~\cite{Leonardo-Software,Real-World}. 
All signals in the database have a 16 bits
resolution and are sampled at 44100 Hz, enabling both good time
resolution and a wide frequency band, which are both necessary to
cover harmonic as well as impulsive sounds. 
The selected sound classes are given in Table~\ref{tab:1}, and they are typical
of surveillance applications. The number of items in each class is
deliberately not equal.

\begin{figure}[t]
  \centering
\includegraphics[height=0.19\textheight, width=\columnwidth]{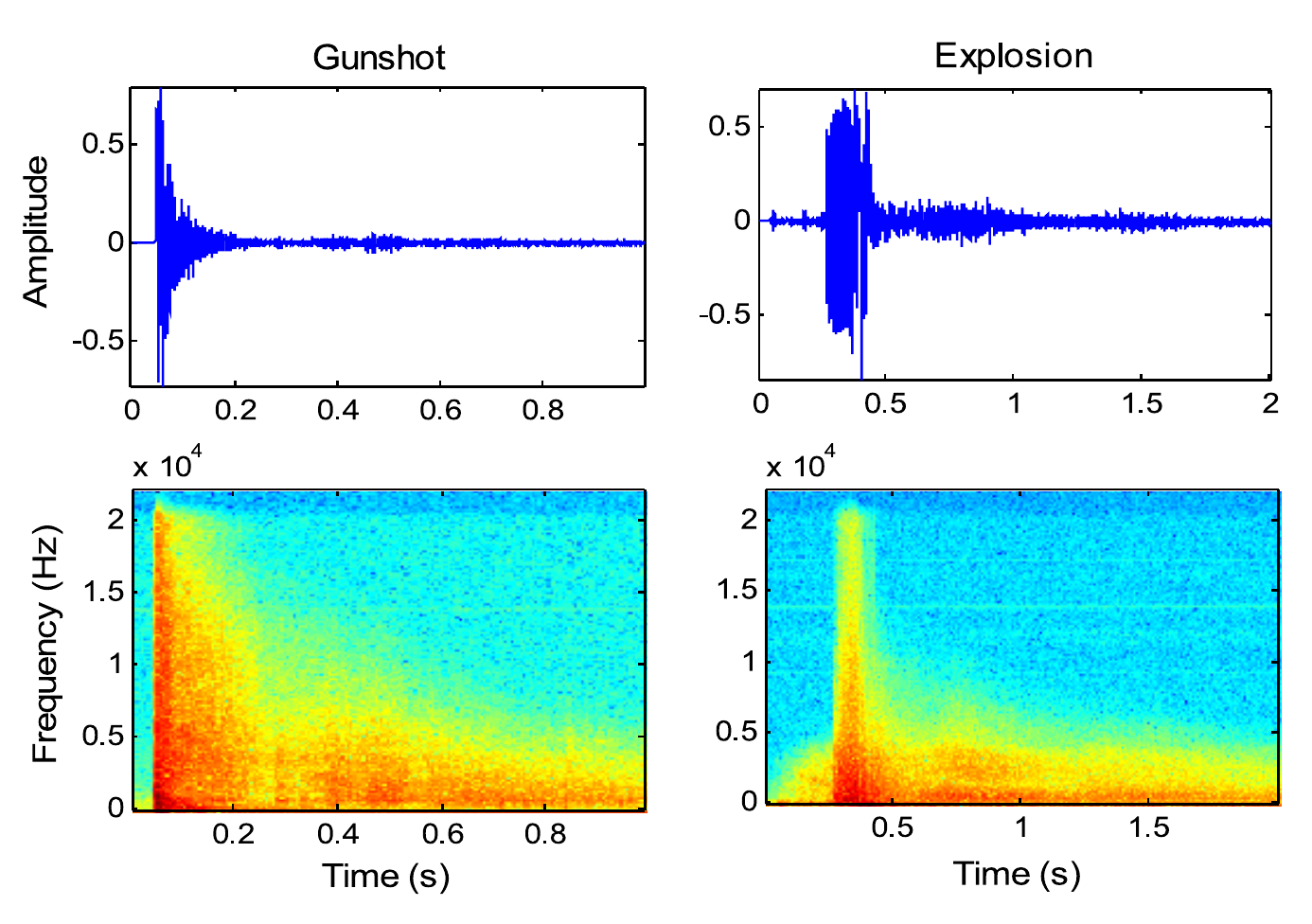}

  \caption{Structural similarities between two different classes.}
\label{similarity}
\end{figure}

\begin{figure}[t]
  \centering
  \includegraphics[height=0.37\textheight, width=\columnwidth]{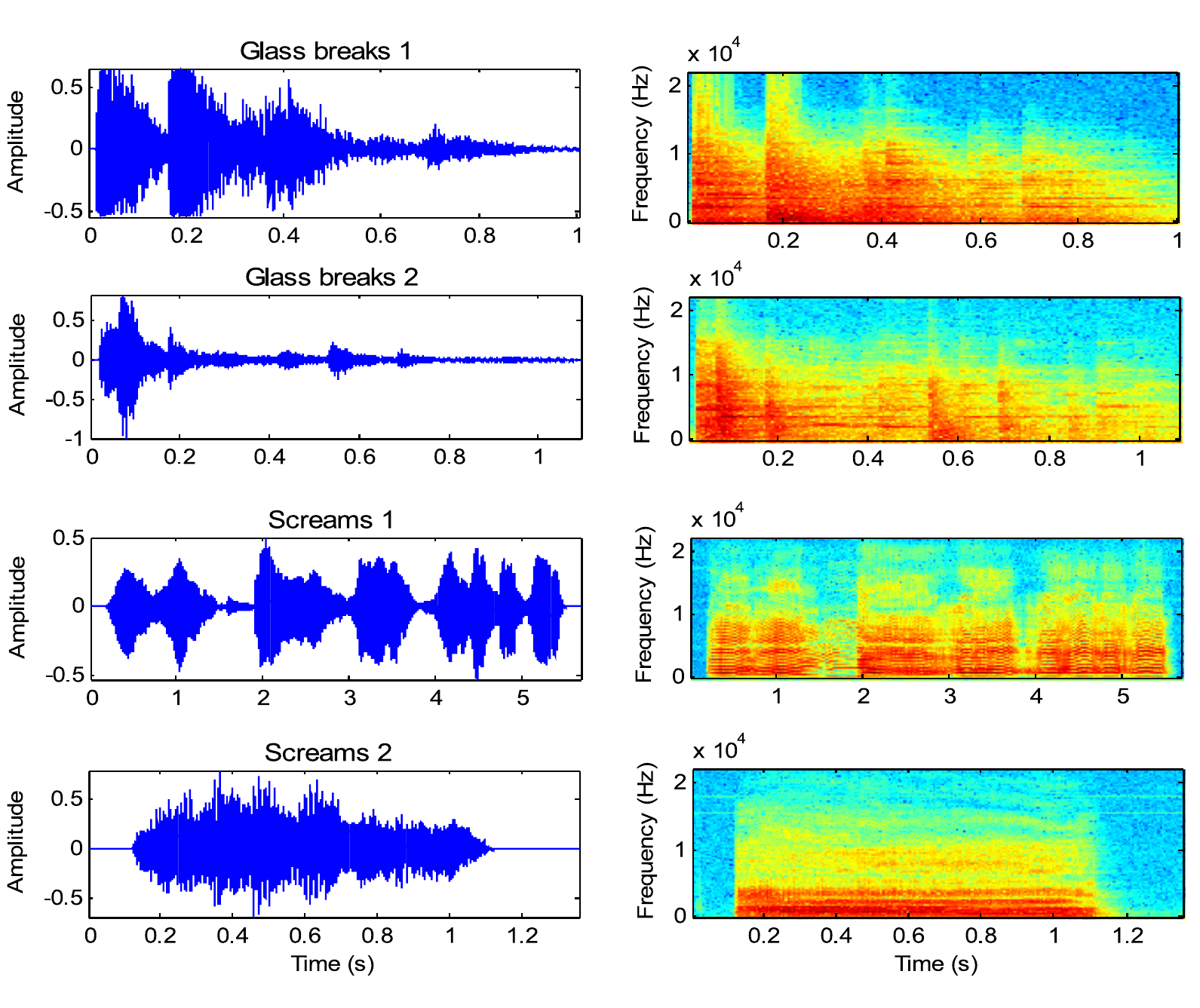}
  \caption{Structural diversity inside the same sound class and between classes.}
\label{structure-temporel}
\end{figure}

Note that this database includes impulsive sounds and harmonic sounds such as phone rings (C6) and children voices (C7). 
These sounds are quite likely to be recorded by a surveillance system. 
Some classes sound very similar to a human listener: in particular, explosions (C4) are pretty similar to gunshots (C2). 
Glass breaking sounds include both bottle and window breaking situations. 
Phone rings are either electronic or mechanic alarms. 
Temporal representations and spectrograms of some sounds are depicted in 
Figure~\ref{similarity} and~\ref{structure-temporel}. 
Power spectra are extracted through the Fast Fourier Transform (FFT) every 10 ms from 25 ms frames. 
They are represented vertically at the corresponding frame indexes. 
The frequency range of interest is between 0 and 22 kHz. 
A lighter shade indicates a higher power value. 
These figures show that in the considered database 
we can have both: (1) many similarities between some sounds belonging to different classes, 
(2) diversities within the same sound class. 
\begin{table}[t]
    \centering
    \caption{Classes of sounds and number of samples in the database used for performance evaluation.\label{tab:1}}
    \scriptsize
\begin{tabular}{D{2cm} D{0.9cm} D{0.5cm} D{0.5cm} D{0.5cm} D{1.3cm}}
  \hline
  Classes & Number &Train & Test & Total  & Duration(s)\tabularnewline
  \hline\hline
Human screams   & C1    & 40    & 25  & 65    &167\tabularnewline
Gunshots        & C2    & 36    & 19  & 55    & 97\tabularnewline
Glass breaking  & C3    & 48    & 25  & 73    &123\tabularnewline
Explosions      & C4    & 41    & 21  & 62    &180\tabularnewline
Door slams      & C5    & 50    & 25  & 75    & 96\tabularnewline
Phone rings     & C6    & 34    & 17  & 51    &107\tabularnewline
Children voices & C7    & 58    & 29  & 87    &140\tabularnewline
Machines        & C8    & 40    & 20  & 60    &184\tabularnewline\hline\hline
  Total         &       & 327   & 181 & 508   &18mn 14s\tabularnewline
  \hline
\end{tabular}
\end{table}
%
%
%
%
%
%
%
%
%
\subsection{Results}

Following~\cite{Rifkin-2004}, the 1-vs-all multi-class classifier is selected in these experiments. 
So we train $N$ (number of classes) different binary classifiers, each one trained to distinguish the data
in a single class from the examples in all remaining classes. We run the $N$ classifiers to classify a new example.~In 
section~\ref{frlsc}, we showed that operator-valued kernels can be used in a classification problem by considering the 
labels $y_i$ to be functions in some function space rather than real values. Similarly to the scalar case, a natural choice 
for $y_i$ would seem to be the Heaviside step function in $L^2([0,1])$ scaled by a real number. 
The used operator valued-kernel is based on the integral operator as defined in~(\ref{kernn}). 
Eigenvalues $\delta_i$ and eigenfunctions $w_i$ associated with this kernel are equal to $\frac{2}{1+\mu_i^2}$ and 
$\mu_i\cos(\mu_ix)+\sin(\mu_ix)$, respectively ; where $\mu_i$ are solutions of 
the equation $\cot \mu = \frac{1}{2}(\mu-\frac{1}{\mu})$.

The adopted sound data processing scheme is the following.~Let $\mathcal{X}$ be the set of training sounds, 
shared in $N$ classes
denoted ${\mathcal{C}}_1,\ldots,{\mathcal{C}}_N$. Each class contains
$m_i$ sounds, $i=1,\ldots,N$. Sound number $j$ in class ${\mathcal{C}}_i$
is denoted ${\mathbf s}_{i,j}$, ($i=1,\ldots,N, j=1,\ldots,m_i$).
The pre-processor converts a recorded acoustic signal
${\mathbf s}_{i,j}$ into a time/frequency localized representation. 
In multivariate methods, this representation is obtained by splitting the signal 
${\mathbf s}_{i,j}$ into $T_{i,j}$ overlapping short frames and computing a vector of features
$z_{t,i,j}$, $t=1,\ldots,T_{i,j}$ which characterize each frame. 
Since the pre-processor is a series of continuous time-localized features, it will be useful 
to take into account the relationships between feature samples along the time axis and consider dependencies 
between features. That is why we use a FDA-based approach in which features representing a sound are 
modeled by functions $z_{i,j}(t)$. 

In this work, Mel Frequency Cepstral Coefficients (MFCCs) are used to describe the spectral shape 
of each signal. These coefficients are obtained using 23 channels Mel filterbank and 
a Hamming analysis window of length $25$ ms with $50\%$ overlap. 
We choose to use $13$ MFCC features and the energy parameter measured along the sound signal. 
So, each sound is characterized by $14$ functional parameters: $13$ cepstral functions and one energy function.
\begin{table}[t]
    \centering
    \scriptsize
    \caption{Confusion Matrix obtained when using the Functional Regularized Least Squares classification algorithm }
    \begin{tabular}{ccccccccc}
        \hline
          & C1& C2 & C3 & C4 & C5 & C6 & C7 & C8  \\
        \hline\hline

        C1 & 100 & 0   & 0  	& 2    & 0 		& 5.3 	& 3.4 & 0  \\
        C2 & 0   & 82  & 0  	& 8    & 0 		& 0 		& 0 	& 0  \\
        C3 & 0   & 14  & 90.9 & 8    & 0    & 0 		& 3.4 & 0 \\
        C4 & 0   & 4   & 0  	& 78   & 0    & 0 		& 0 	& 0  \\
        C5 & 0   & 0   & 0  	& 1 	 & 89.47& 0 		& 6.8 & 0  \\
        C6 & 0   & 0   & 0  	& 0 	 & 10.53& 94.7  & 0 	& 0 \\
        C7 & 0   & 0   & 0  	& 0 	 & 0 		& 0 		& 86.4& 0  \\
        C8 & 0   & 0   & 9.1 	& 3	   & 0	  & 0 		& 0 	& 100  \\
        \hline\hline
    \multicolumn{9}{c} {\emph{Total Recognition Rate = 90.18\%}} \\

    \hline
    \end{tabular}
    \label{tab1}
\end{table}
\begin{table}[t]
    \centering
        \scriptsize
    \caption{Confusion Matrix obtained when using the Regularized Least Squares Classification (RLSC) algorithm}
    \begin{tabular}{ccccccccc}
        \hline
          & C1& C2 & C3 & C4 & C5 & C6 & C7 & C8 \\
        \hline\hline

        C1 & 92  & 4     & 4.76  & 0 			& 5.27 	& 11.3 	& 6.89 		& 0  \\
        C2 & 0   & 52    & 0  	 & 14     & 0		 	& 2.7 	& 0 		  & 0  \\
        C3 & 0   & 20    & 76.2  & 0 		& 0    	& 0 		& 17.24		& 5 \\
        C4 & 0   & 16    & 0     & 66     & 0    	& 0 		& 0 		  & 0  \\
        C5 & 4   & 8     & 0  	 & 4  		& 84.21 & 0 		& 6.8		  & 0  \\
        C6 & 4   & 0     & 0     & 0 			& 10.52 & 86  	& 0 		  & 0 \\
        C7 & 0   & 0     & 0     & 8 			& 0    	& 0 		& 69.07 	& 0  \\
        C8 & 0   & 0     & 19.04 & 8 			& 0    	& 0 		& 0 		  & 95  \\
        \hline\hline
    \multicolumn{9}{c} {\emph{Total Recognition Rate = 77.56\%}} \\

    \hline
    \end{tabular}
    \label{tab2}
\end{table}

Performance of the Functional RLSC based classifier is compared to the results obtained by the
RLSC algorithm, see Table~\ref{tab1} and~\ref{tab2}. 
The performance is measured as the percentage number of sounds correctly recognized and it is
given by $(\mathrm{W_r}/\mathrm{T_n})\times100\%$, where $\mathrm{W_r}$ is the number of well recognized sounds 
and $\mathrm{T_n}$ is the total number of sounds to be recognized. 
The use of the Functional RLSC is fully justified by the results presented
here, as it yields consistently lower error rates and a high
classification accuracy for the major part of the sound classes. 
%

%
%
%
%
%
\section{Conclusion}
\label{conc}

This paper has put forward the idea that by viewing operator-valued kernels from a feature map perspective, 
we can design more general kernel methods well-suited for complex data. Based on this, we have extended the regularized 
least squares classification algorithm to functional data analysis contexts where input data are 
real-valued functions rather than finite dimensional vectors. 
Through experiments on sound recognition, we have shown that the proposed approach is efficient and improves the 
classical RLSC method in a sound classification dataset. 
Further investigations will consider larger classes of 
operator-valued kernels. 
It would also be interesting to study learning methods for choosing the operator-valued kernel.

\section*{Acknowledgments}
This work was supported by Ministry of Higher Education and Research, Nord-Pas-de-Calais Regional 
Council and FEDER through the `Contrat de Projets Etat Region (CPER) 2007-2013'.~H.K. is 
supported by Junior Researcher Contract No.~4297 from the Nord-Pas-de-Calais 
region.~A.R. is supported by the PASCAL2 Network of Excellence, ICT-216886, ANR Project 
ASAP ANR-09-EMER-001 and the INRIA ARC MABI.

\bibliography{HK_ICML2011}
\bibliographystyle{icml2011}
\end{document}